\documentclass[10pt,twocolumn,letterpaper]{article}

\usepackage{cvpr}              % To produce the CAMERA-READY version
\usepackage[dvipsnames]{xcolor}

\definecolor{cvprblue}{rgb}{0.21,0.49,0.74}
\usepackage[pagebackref,breaklinks,colorlinks,citecolor=cvprblue]{hyperref}

\usepackage{graphicx}
\usepackage{cuted}
\usepackage{caption}

\usepackage{float} % Allows for [H] option

\def\paperID{378} % *** Enter the Paper ID here
\def\confName{3DV\xspace}
\def\confYear{2025\xspace}

\title{LEMON: Localized Editing with Mesh Optimization and Neural Shaders}

\author{Furkan Mert Algan  \quad Umut Yazgan \quad Driton Salihu  \quad Cem Eteke  \quad Eckehard Steinbach\\
Technical University of Munich\\
Chair of Media Technology and Munich Institute of Robotics and Machine Intelligence (MIRMI)
\\School of Computation, Information and Technology
Department of Computer Engineering
\\
{\tt\small \{fmert.algan, umut.yazgan, driton.salihu, cem.eteke, eckehard.steinbach\}@tum.de }
}

\begin{document}

% \twocolumn[
% \maketitle
% \begin{figure*}[!t]
%   \centering
%   \includegraphics[width=0.7\textwidth]{figures/intro_fig.pdf}
%   \caption{
% Lorem ipsum dolor sit amet, consectetur adipiscing elit, sed do eiusmod tempor incididunt ut labore et dolore magna aliqua. Ut enim ad minim veniam, quis nostrud exercitation ullamco laboris nisi ut aliquip ex ea commodo consequat.}
%   \label{fig:opening}
% \end{figure*}

% ]

% \begin{figure*}[!t] % [h] helps to keep the figure in place
%   \centering
%   \includegraphics[width=0.7\textwidth]{figures/intro_fig.pdf} % adjust width as needed
%   \caption{
% We propose LEMON, a polygonal mesh editing method that takes multi-view images and user-provided text instructions as input and edits the mesh while preserving the geometric characteristics of the original mesh}
%   \label{fig:opening} % label for referencing
% \end{figure*}
\maketitle

\begin{strip}
    \centering
    \includegraphics[width=1.9\columnwidth]{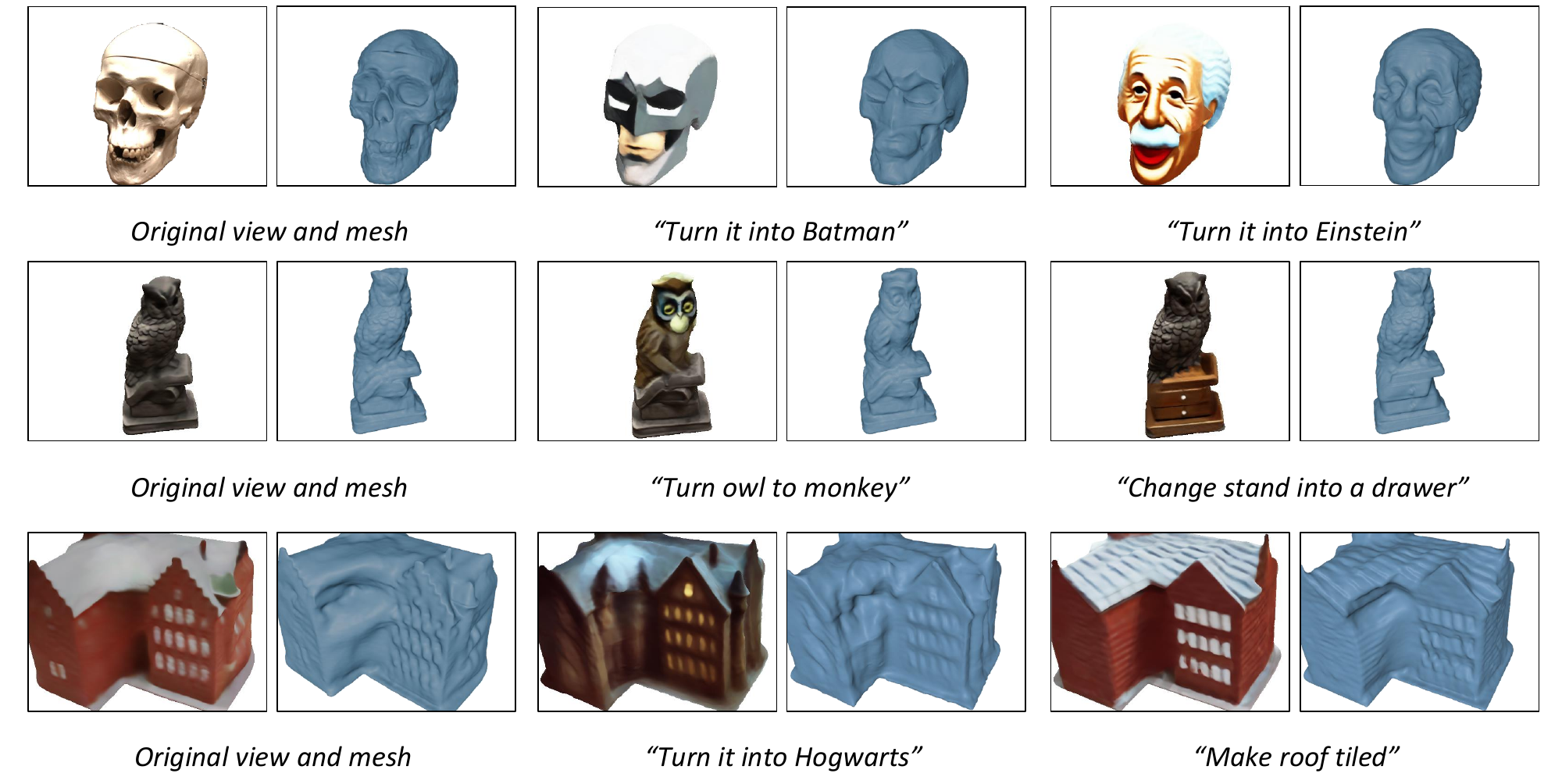}
    \captionof{figure}{We propose LEMON, a polygonal mesh editing method that takes multi-view images and user-provided text instructions as input and edits the mesh while preserving the geometric characteristics of the original mesh. Our method localizes accordingly to given instruction only changes the important parts of the mesh and provides a neural shader for the novel view.}
  \label{fig:opening}
\end{strip}

\begin{abstract}
In practical use cases, polygonal mesh editing can be faster than generating new ones, but it can still be challenging and time-consuming for users. Existing solutions for this problem tend to focus on a single task, either geometry or novel view synthesis, which often leads to disjointed results between the mesh and view. In this work, we propose LEMON, a mesh editing pipeline that combines neural deferred shading with localized mesh optimization. Our approach begins by identifying the most important vertices in the mesh for editing, utilizing a segmentation model to focus on these key regions. Given multi-view images of an object, we optimize a neural shader and a polygonal mesh while extracting the normal map and the rendered image from each view. By using these outputs as conditioning data, we edit the input images with a text-to-image diffusion model and iteratively update our dataset while deforming the mesh. This process results in a polygonal mesh that is edited according to the given text instruction, preserving the geometric characteristics of the initial mesh while focusing on the most significant areas. We evaluate our pipeline using the DTU dataset, demonstrating that it generates finely-edited meshes more rapidly than the current state-of-the-art methods. We include our code and additional results in the supplementary material.

\end{abstract}

%-------------------------------------------------------------------------

\section{Introduction}
\label{sec:intro}
  3D object representations can now be obtained with relative ease through neural rendering methods such as NeRFs~\cite{mildenhall2021nerf} and Gaussian Splatting~\cite{gaussiansplatting}. By using multi-view pictures from calibrated cameras and optimizing a neural network, 3D object representations can be generated efficiently. Yet, if we want to edit a representation while retaining its unique characteristics instead of generating a new one, manual editing is often the only viable option.

Although manual editing methods offer greater control to the user, this approach can be cumbersome and time-consuming because the user has to process each piece of data themselves. Instruct-NeRF2NeRF~\cite{Haque_2023_ICCV} and more recently GaussianEditor~\cite{chen2024gaussianeditor} deliver impressive results in editing 3D scenes from multi-view images based on user-provided text prompts, but it also carries the inherent limitations of novel view methods. A significant drawback is the challenge of extracting meshes efficiently, represent scenes as continuous volumetric functions,and Gaussian Splatting represents them using Gaussian distributions. This limits their use in many cases where polygonal meshes are needed, as it can be challenging to maintain the characteristics of the edited object while incorporating new details.

We propose LEMON, a localized mesh editing method that deforms polygonal meshes based on given text instructions while preserving 3D consistency. Given multi-view images of the mesh, we use CLIPSeg~\cite{CLIPSeg} to generate a vertex scores to determine important vertices in the given context. Afterwards, we use normal maps and shaded images of the mesh obtained through neural deferred shading~\cite{nds} as conditions of the ControlNet~\cite{controlnet} to achieve 3D consistent images. Based on vertex scores, we mask edited images to localize our modifications. We iteratively update our multi-view image dataset with modified pictures and deform meshes based on these edits and vertex scores. Through this process, we ensure that the resulting meshes reflect the text-based instructions while maintaining the initial 3D structure.

\section{Related Work}
\label{sec:relatedwork}

\textbf{Neural Rendering:}  As deep learning techniques gain prominence in both computer vision and graphics, neural rendering~\cite{tewari2022advances} plays an important role in 3D reconstruction. One of the most popular techniques, neural radiance fields(NeRF)~\cite{mildenhall2021nerf}, are a type of volumetric scene representation based on a continuous volumetric function parameterized by a multilayer perceptron(MLP). NeRF can produce photorealistic renderings but is computationally expensive and slow. Additionally, most NeRF methods focus on view synthesis and rely on other techniques for surface extraction~\cite{kazhdan2006poisson}, which often results in less accurate meshes. In recent years, Gaussian Splatting ~\cite{gaussiansplatting} has emerged as an efficient alternative, offering faster rendering by representing scenes with a collection of Gaussian kernels, though it also struggles with precise surface reconstruction. Recent works, such as ~\cite{guedon2024sugar}, show that it is possible to extract meshes from Gaussians. However, rather than being direct polygonal mesh editing, it is more of a post-processing step that adds extra time to the editing process. By contrast, we use neural deferred shading(NDS)~\cite{nds}, which integrates traditional mesh deformation with a neural shading pipeline, enabling more precise and detailed 3D reconstructions from multi-view images.
% More recently Kerbl et al.~\cite{gaussiansplatting} introduced 3D Gaussian Splatting

\textbf{Diffusion Models:} Recent breakthroughs in diffusion models provide a more flexible approach to creating images from text or other conditioning data. Denoising Diffusion Implicit Models (DDIM)~\cite{song2020denoising} iteratively remove noise from an initial noisy input to generate samples, while Contrastive Language–Image Pre-training (CLIP)~\cite{CLIP} guides image generation based on text prompts. Stable Diffusion~\cite{stablediffusion} is known for its high-quality outputs while InstructPix2Pix~\cite{instructpix2pix} extends this concept by using text prompts to edit existing images.  However, expressing complex layouts through text prompts alone can be challenging in text-to-image models. We choose  ControlNet~\cite{controlnet} because it addresses this issue by incorporating spatial conditioning controls, such as normal maps, which helps maintain the geometric characteristics of the object.

%\textbf{Mesh Generation:} 3D Advancements in 2D image generation have also contributed to text-to-3D generation techniques. CLIP-Mesh ~\cite{clipmesh} introduced a text-driven approach to 3D content generation by using a pre-trained CLIP model. DreamFusion ~\cite{poole2022dreamfusion} improved this by optimizing a NeRF based on a text-to-image diffusion model. Magic3D ~\cite{lin2023magic3d} further improved DreamFusion's resolution through a coarse-to-fine optimization approach. Despite these innovations, all of these methods focus on implementing a mesh from scratch and lack support for editing already existing meshes, making their variation dependent on their respective pre-trained diffusion models.

\textbf{Mesh Editing:} Despite the extensive research in mesh generation~\cite{poole2022dreamfusion, clipmesh, lin2023magic3d}, there are relatively few studies focusing on 3D model editing, and these typically do not involve polygonal mesh editing. Instruct-NeRF2NeRF~\cite{Haque_2023_ICCV} introduced a text-based editing technique for NeRF scenes, where an image-conditioned diffusion model~\cite{instructpix2pix} generates new images based on a NeRF representation of a scene and the images used to construct it, which are then used to iteratively update the training dataset, guiding the NeRF to converge to the edited version. However, this technique relies on an implicit representation rather than directly manipulating the surface geometry, which limits its applicability to polygonal mesh editing. GaussianEditor~\cite{chen2024gaussianeditor} follows a similar logic as Instruct-NeRF2NeRF) but applies it to Gaussian splatting, enabling text-based editing of Gaussian representations. Yet, for precise mesh extraction and reconstruction, additional pipelines like SuGaR ~\cite{guedon2024sugar} are necessary to efficiently convert Gaussian splats into accurate 3D meshes. Inspired by~\cite{aigerman2022neural}, TextDeformer~\cite{gao2023textdeformer} is capable of creating global deformations on an input mesh based on a text prompt. However direct manipulation of meshes through Jacobians is computationally costly and they do not provide any color information about the edited mesh. 
In contrast, our method deforms a polygonal mesh using a fast neural shader combined with a diffusion model, giving the deformable mesh additional context about color and specularity. %thereby preserving the characteristics of the initial mesh during editing.

\begin{figure*}[t]
  \centering
  \includegraphics[width=2.0\columnwidth]{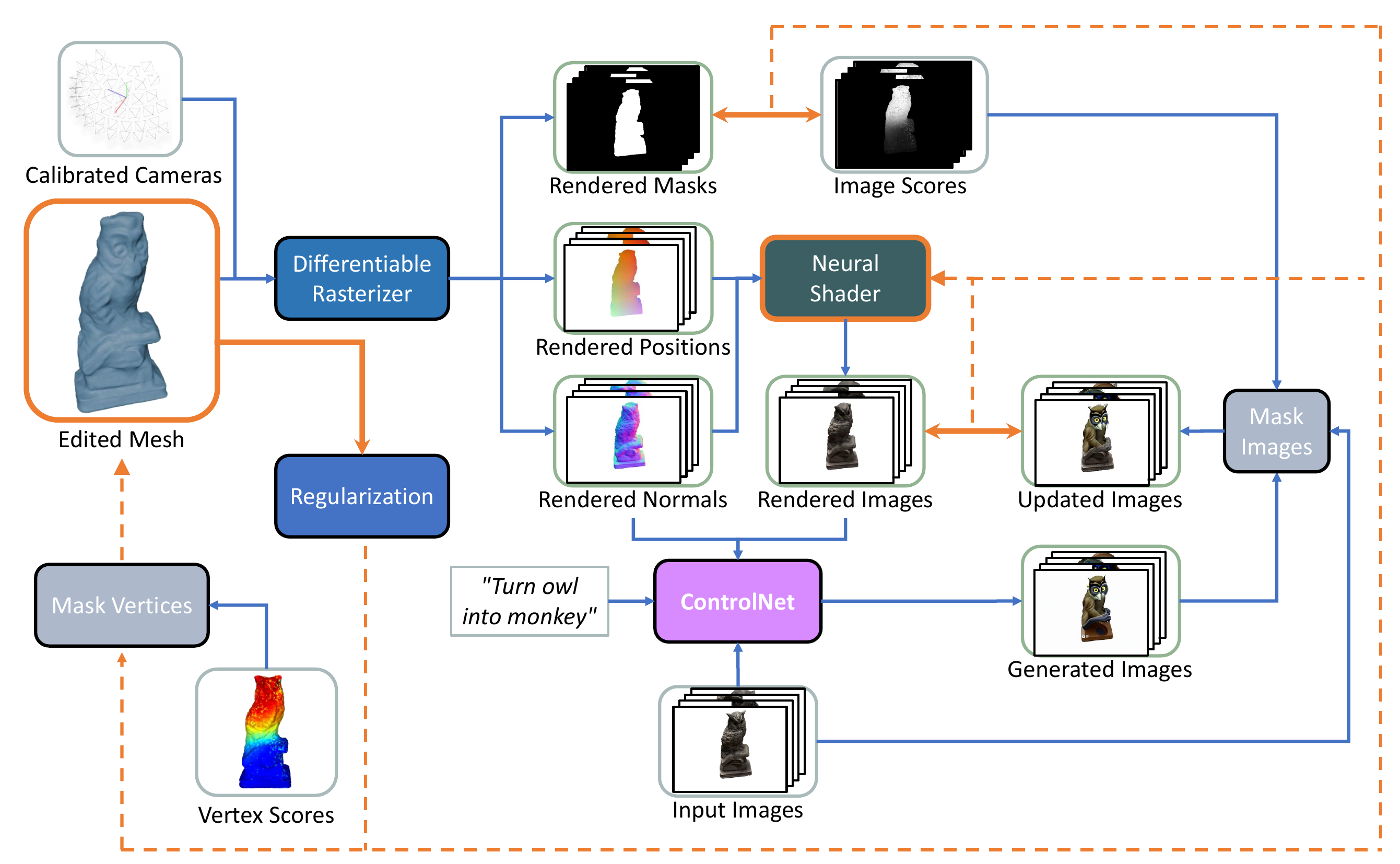} 
  \caption{Pipeline of LEMON: We complement a multi-view mesh reconstruction model with a text-to-image model using localized features. After gathering vertex scores in our pre-processing step, we begin our editing process. Every $d$ iterations new images are generated by the ControlNet~\cite{controlnet}, based on the prompt. The initial noise calculation of diffusion model is derived from a weighted sum of input images and rendered images, while it is conditioned on rendered normals and images of the mesh. The generated images are masked, and the masked regions are overlaid onto the original images, creating modified versions that are then used to update the dataset. Using vertex scores as a mask on the mesh, we update only the subset of vertices that is relevant to the prompt. By continuously updating the dataset with edited images, we deform the mesh to align with the user's request.}
  \label{fig:main-figure} % label for referencing
\end{figure*}

\section{Method}
\label{sec:method}
We propose LEMON, a fast and lightweight method for editing polygonal meshes by combining neural deferred shading and text-to-image models. We take a set of multi-view $m$ images, \(\mathcal{I} = \{I^1, \ldots, I^m\}\), from calibrated cameras, along with corresponding masks of the interested zone, \(\mathcal{M} = \{M^1, \ldots, M^m\}\), and the camera parameters, as input. As in~\cite{nds}, we represent the surfaces as a triangle mesh, \(\mathcal{G} = (\mathcal{V}, \mathcal{E}, \mathcal{F})\), consisting of vertices \(\mathcal{V}\), edges \(\mathcal{E}\), and faces \(\mathcal{F}\). Additionally, we take a text prompt \(T\) for the editing instruction. The output of our method is an edited version of the initial mesh and a neural shader, based on the user-provided instruction. In this section, we first provide background information on the pipelines we've used; then we explain how our method integrates them.

\subsection{Background}
\textbf{Neural Deferred Shading:} NDS~\cite{nds} is an analysis-by-synthesis mesh reconstruction method that optimizes a mesh and a neural shader simultaneously by using calibrated images and their corresponding masks as its input. If the initial mesh is not provided the optimization process begins with a mesh that is derived from the masks and resembles a visual hull~\cite{laurentini1994visual}. Starting with a coarsely triangulated mesh, its resolution is gradually increased as optimization proceeds. In each upsampling iteration the surface is remeshed  reducing the average edge length by half~\cite{botsch2004remeshing}.

In the first step, the mesh is rasterized using a differentiable renderer~\cite{nvdiffrast}, which provides triangle indices and barycentric coordinates for each pixel. By interpolating this information, a geometry buffer (g-buffer) is generated, containing per-pixel positions, normals, and mask information. In second step g-buffer is processed by a learned shader, a MLP which resulting in an RGB color:

\begin{equation}
f_{\theta} (x, n, {\omega}_o) \in [0, 1]^3
\label{eq:learnedshader}
\end{equation}

where $\theta$ represents the learnable parameters, $x \in {\mathbb{R}}^3$ is the position, $n \in {\mathbb{R}}^3$ denotes the normal and $\omega_o \in {\mathbb{R}}^3$ is the view direction relative to the center of the camera. 

The neural shader and the initial mesh is optimized based on an objective function that balances the rendered appearance of mesh and geometric characteristics of the mesh:

\begin{equation}
\underset{V, \theta}{\arg \min } L_{\text {appearance }}(\mathcal{G}, \theta ; \mathcal{I}, \mathcal{M})+L_{\text {geometry }}(\mathcal{G})
\label{eq:ndsoptim}
\end{equation}

where $L_{\text {appearance }}$ consists of a shading loss which computes distance between rendered image $\mathcal{\tilde{I}}$ and input image $\mathcal{I}$ and a mask loss, which computes distance between rendered mask $\mathcal{\tilde{M}}$ and input mask $\mathcal{{M}}$. $L_{\text {geometry }}$ ensures to avoid undesired vertex configurations while deforming the mesh by minimizing the distance between a vertex and the average position of the neighbors and computing cosine similarity between neighboring face normals. Because mesh deformation and shader optimization happen simultaneously, NDS can preserve the geometric characteristics of opaque objects, even with varying materials and illumination, making it an ideal candidate for use in mesh editing.

\textbf{ControlNet:} Denoising diffusion models~\cite{croitoru2023diffusion} generate data by incrementally removing noise, with U-Net architecture~\cite{unet} providing both local and global context during the denoising process. Text-to-image diffusion models like Instruct-Pix2Pix~\cite{instructpix2pix} produce high-quality images from text-based instructions, encoding text into latent vectors using pretrained language models like CLIP~\cite{CLIP}. However, these models often lack control over specific conditions and rely on broad assumptions about user preferences. Moreover, training new models can be time-consuming and burdensome, particularly with large datasets. 

ControlNet~\cite{controlnet} freezes the original parameters of a diffusion model and creates a trainable copy of its encoding layers, which are then trained on specific conditions. This method maintains the original diffusion model's quality and functionality, while allowing for more control through defined conditions. In our case, extra conditions like normal maps provide valuable supervision on geometry of the mesh, which is why we chose ControlNet as our diffusion model in our mesh editing pipeline.

\subsection{LEMON}
Starting with an initial polygonal mesh (along with a corresponding dataset of calibrated images and their camera parameters), our method combines a diffusion model with a neural deferred shading pipeline to deform the mesh based on given textual instruction.  If the initial mesh is not provided, it is created using standard neural deferred shading pipeline.~\cite{nds}

We closely follow Instruct-NeRF2NeRF~\cite{Haque_2023_ICCV}'s editing process albeit with some differences. First we perform a pre-processing step to determine important regions of the mesh based on given instruction. During our editing process, we store another set of ground truth multi-view images, which we denote as $\mathcal{{I}}^v$. After a certain number of iterations \(d\) we update our training dataset by replacing images with modified ones. Sometime later our mesh deforms into the desired edited version. Overview of our pipeline is given in Figure \ref{fig:main-figure}.

\begin{figure}[t] % or [htb]
    \centering
    \includegraphics[width=\columnwidth]{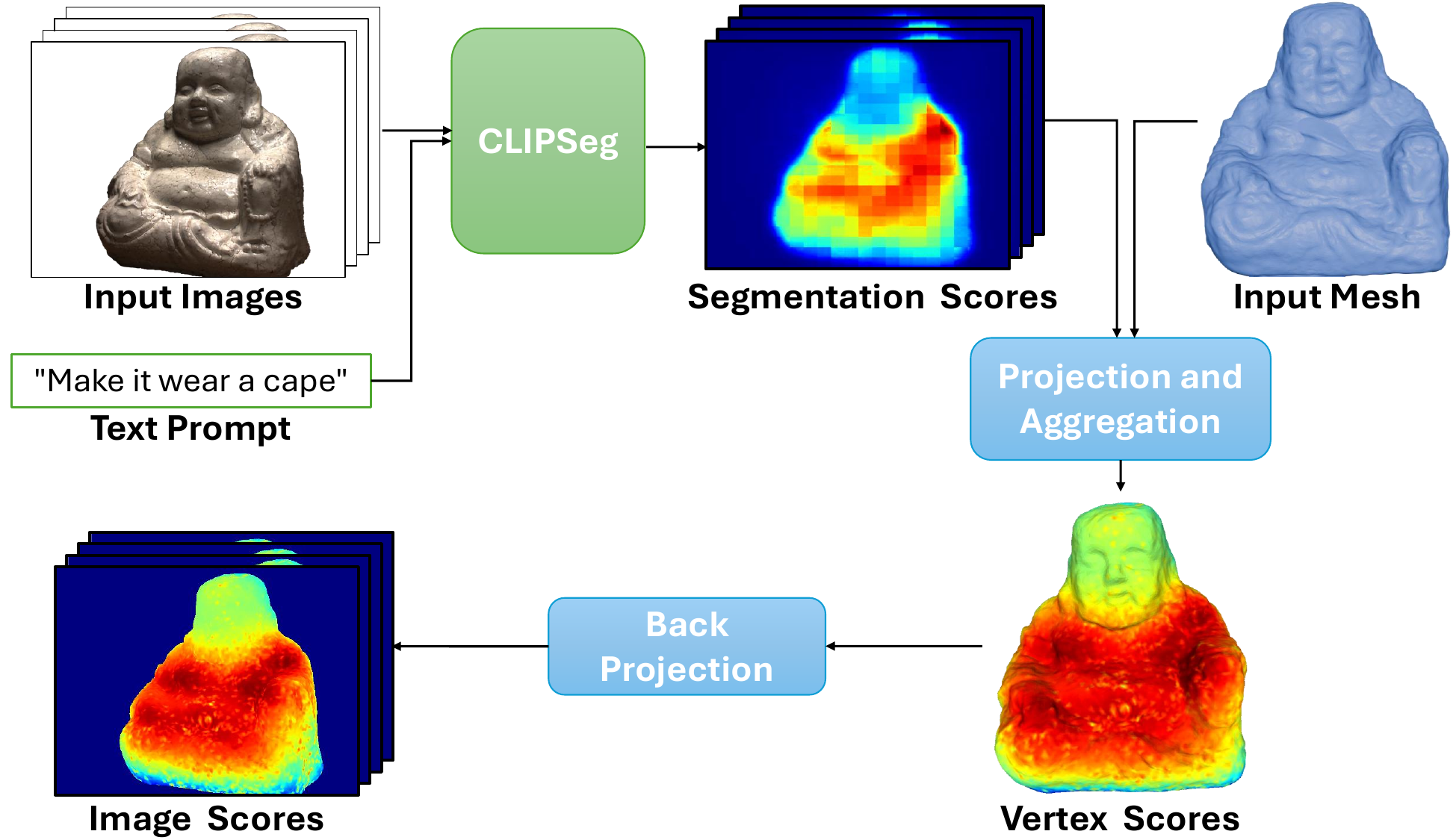}
    \caption{Image and vertex scoring process. Using CLIPSeg~\cite{CLIPSeg} we segment most important parts of the mesh given instruction}
    \label{fig:segmentation-projection}
\end{figure}

\textbf{Vertex and Image Scores:} For the first step of our pre-processing, we generate segmentation masks for every calibrated image using the provided text prompt. To keep it simple for the user and ensure compatibility with our image editing process using CLIP~\cite{CLIP}, we have selected CLIPSeg~\cite{CLIPSeg} to generate segmentation scores. These scores highlight the regions that are relevant to the given prompt as seen in Figure \ref{fig:segmentation-projection}.

To assign these scores to the vertices, we use NvDiffrast ~\cite{nvdiffrast} to render the mesh from given viewpoints and project segmentation scores onto the mesh surface. Each vertex's score is then calculated by averaging the projected scores from all viewpoints. This aggregated scoring system makes the segmentation scores more accurate based on the geometry, effectively representing each vertex's importance in depicting the given prompt's feature distribution across the 3D mesh. After calculating the segmentation scores for each vertex, we project these scores back onto the image viewpoints. This process ensures that the segmentation scores are accurately reflected in the 2D image space.

In order to provide the user with more control over how much the mesh changes based on the context, the threshold $\tau \in [0, 1]$ is also taken as an input. By keeping scores larger than $\tau$, we gather new image masks \(\mathcal{\widetilde{M}} = \{\widetilde{M}^1, \ldots, \widetilde{M}^m\}\) and a subset of vertices \(\mathcal{\widetilde{V}}\subset\mathcal{V}\) whose scores are higher than $\tau$. We optimize only this subset while editing the mesh. We are doing these processes to ensure that changes occur only in the most important regions, localized according to the context of the instruction.

\begin{figure}
  \centering
  \includegraphics[width=1\columnwidth ]{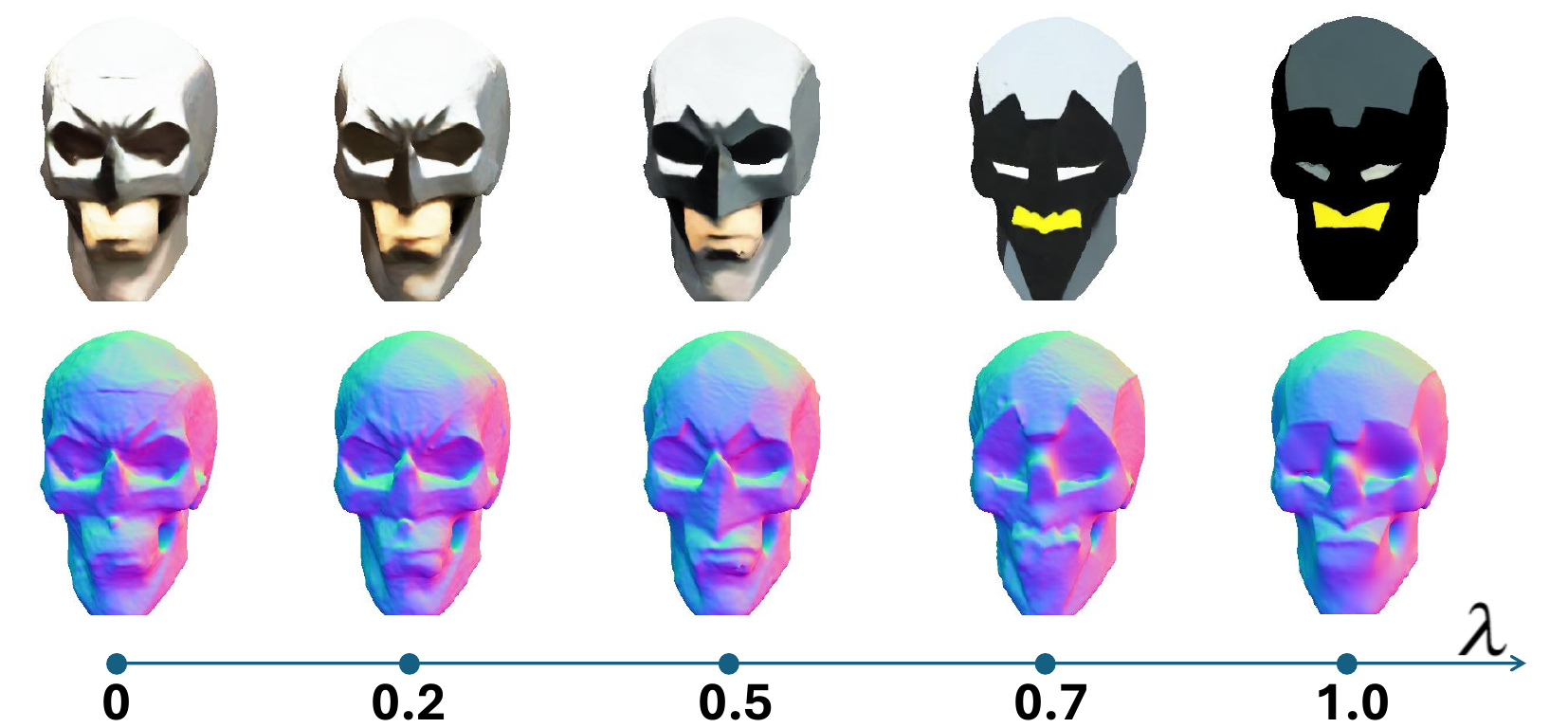} % adjust width as needed
\caption{The effect of the latent weight hyperparameter $\lambda$ on editing of the skull object from ~\cite{DTU} for the "Turn it into Batman" prompt. Top of the skull has very bright shading, but the prompt requires the object to be darker. When $\lambda$ is set to 0, only the ground truth image is used for the initial noise calculation, resulting in the lines in the skull to stay. If $\lambda$ is too high, the rendered image may diverge to darker tones, leading to unintended edits.}
\label{fig:man_Results} % optional label for cross-references
\end{figure}

\textbf{Editing Image:} For the editing process, we use a pre-trained variant of ControlNet~\cite{controlnet} with two modules, one fine-tuned on Instruct-Pix2Pix~\cite{instructpix2pix} images and the other on normal maps. For calculation of the initial noise $z_0$, Instruct-NeRF2NeRF follows the approach of SDEdit~\cite{meng2021sdedit} where the rendered image of the current global 3D model plays a role in the output of the diffusion model. Instead of using only the rendered image, we use a weighted sum of the rendered image $\mathcal{\tilde{I}}^v$ and the original input image $\mathcal{{I}}^v$ from a given viewpoint \(v\).

\begin{equation}
    z_0 = \sqrt{\hat{\alpha_{0}}} (\lambda \mathcal{E}(\mathcal{\tilde{I}}^v) + ( 1 - \lambda) \mathcal{E}(\mathcal{{I}}^v))+\sqrt{1-\hat{\alpha_{0}}} \epsilon
\label{eq:initialnoise}
\end{equation} %TODO should we seperate this as two functions

where $\epsilon \sim \mathcal{N}(0,1)$, $\hat{\alpha_{0}}$ is the noise scheduling factor at timestep 0 and $\mathcal{E}$ is the CLIP image encoder. Hyperparameter $\lambda$ is the latent weight used to control the proportion of $\mathcal{\tilde{I}}^v$ and $\mathcal{{I}}^v$ latents. In the case of $\lambda = 1$, the initial noise calculation is the same as Instruct-NeRF2NeRF. Since our focus is on editing rather than mesh generation, we want our edited mesh to be constrained on the initial input pictures as well. However in order to be restrained not too much on initial mesh we also add rendered images to add some variance to the input noise. This allows the diffusion model to not diverge too much to the dark and bright images as seen in Figure \ref{fig:man_Results}.

As conditions for the diffusion model, we include the encoded text prompt $c_T$, generated by~\cite{CLIP} to guide our pipeline for editing. For the normal map module, we provide the normal map $n_i$ of the g-buffer from a given viewpoint \(v\) as conditioning input. This leads to generated images that are more geometrically consistent, directly influenced by the optimized mesh \(\mathcal{G}\). Instead of using ground-truth images for conditions as in Instruct-NeRF2NeRF, our Instruct-Pix2Pix module uses rendered images $\mathcal{\tilde{I}}^v$ generated by \(f_{\theta}\) as its conditioning input. This results in generated images with greater variability, directly influenced by the optimized neural shader \(f_{\theta}\). In a sense our neural deferred shading pipeline and our conditioning inputs optimize each other, resulting in more consistent image generation.

\textbf{Optimizing Process:} We adapt the iterative dataset update of Instruct-NeRF2NeRF to our neural deferred shading pipeline. After the generation of the initial mesh, every \(d\) iteration we create a modified input image using our diffusion model and change the input image of the current iteration with the modified image. We gradually optimize vertices of the mesh and neural shader based on new input images. The equation below demonstrates the image update process.

\begin{equation}
    \mathcal{{{I}}}_{i+1}^v \leftarrow U_\theta\left(\mathcal{I}_0^v, \mathcal{{\tilde{I}}}_i^v, t ; \mathcal{{\tilde{I}}}_i^v, {n_i}^v, c_T\right)\odot{\widetilde{M}^v} + \mathcal{I}_0^v\odot (1-{\widetilde{M}^v})
\end{equation}

where \( t \) is the noise level, randomly selected between \([t_{min}, t_{max}]\) and \( U_{\theta} \) is the DDIM~\cite{song2020denoising} sampling process with a set number of intermediate steps \( s \) between the initial timestep \( t \) and $0$.  As in NeRFs, this approach ensures consistent mesh transformations, providing desired modifications while preserving the mesh's structural integrity.

To localize our editing, we use \(\mathcal{\widetilde{M}}\) to mask the important regions of the generated images. By overlaying the masked generated images onto the original input images, we ensure that modifications are consistently applied only to the relevant areas during the dataset update process.

Unlike NDS, which focuses on all the vertices of the mesh, we only optimize editing subset $\mathcal{\widetilde{V}}$. This approach allows the mesh to focus on editing only the parts that are relevant to the text input, avoiding unnecessary changes and extra computation. By updating only the relevant vertices, we maintain the overall structure and integrity of the original mesh, resulting in more contextually appropriate edits in geometry.

 % use [h] to keep the figure in place or [t], [b], [p] for top, bottom, or a separate page
\begin{figure*}
  \centering
  \includegraphics[width=2\columnwidth, ]{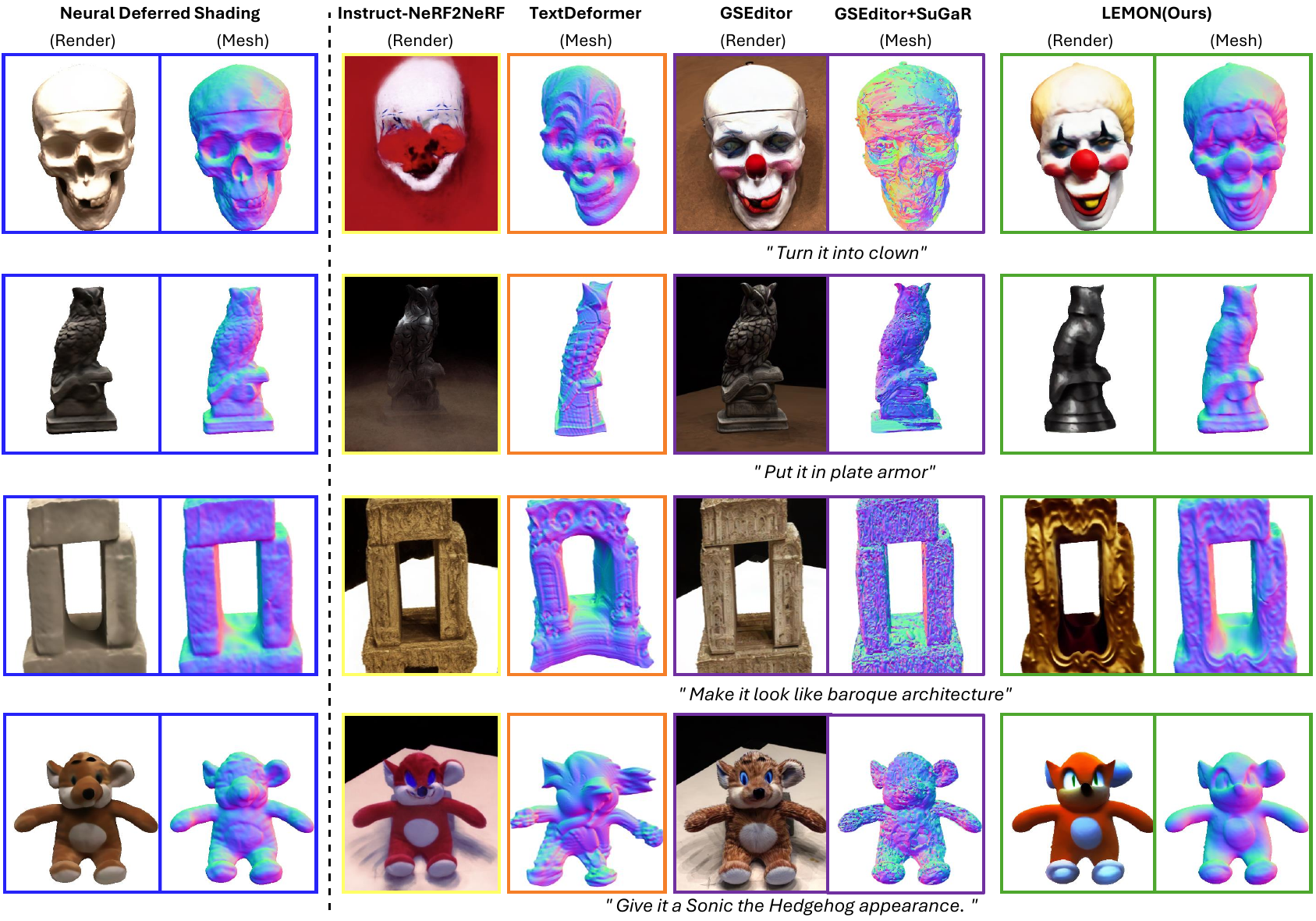} % adjust width as needed
\caption{Editing results on the DTU dataset~\cite{DTU}. \textcolor{blue}{Blue boxes} represent the initial mesh and shader reconstructed by neural deferred shading~\cite{nds}, providing a baseline.  \textcolor{orange}{Orange boxes} show the edited mesh results from TextDeformer while  \textcolor{yellow}{yellow boxes} represent the edited views from Instruct-NeRFNeRF. \textcolor{violet}{Violet boxes} represent renderings from GaussianEditor and their meshes extracted by SuGaR. LEMON achieves great results in both rendering and polygonal mesh quality.}
\label{fig:dtu_results}
\end{figure*}

\section{Experiments}

\begin{figure}
  \centering
  \includegraphics[width=\columnwidth, height=0.5\columnwidth ]{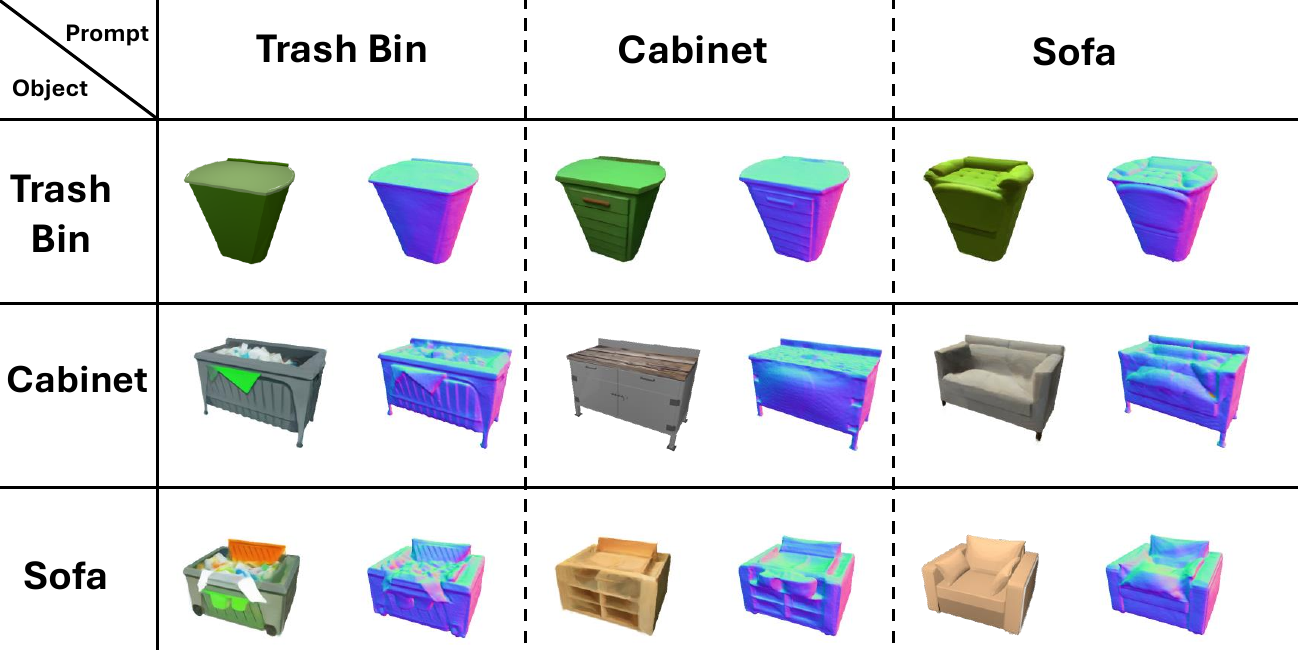} % adjust width as needed
\caption{ShapeNet~\cite{shapenet2015} transformation results. Diagonals corresponds to initial mesh and rendering of the ShapeNet object.}
\label{fig:shapenet_small} % optional label for cross-references
\end{figure}

We evaluate our method both qualitatively and quantitatively on nine objects from the DTU multi-view dataset~\cite{DTU}, using materials from earlier works~\cite{nds, idr}, based on three text prompts. For our method, we first reconstruct the object by following the approach in~\cite{nds}, then proceed with editing. Since there isn't any specific mesh editing method with a shader for direct comparison, we have selected a mesh deformation and novel view synthesis techniques for comparison. For rendering, we selected the Instruct-NeRF2NeRF~\cite{Haque_2023_ICCV} model from Nerfstudio~\cite{tancik2023nerfstudio} and GSEditor~\cite{chen2024gaussianeditor}. Since NeRF and Gaussian Splatting are designed for novel view synthesis rather than surface reconstruction, they are less suitable for mesh-based comparisons. However, recent works~\cite{guedon2024sugar} show that high-quality meshes can be extracted from Gaussian Splatting. Therefore, we compared the renderings of Instruct-NeRF2NeRF and GSEditor, as well as the meshes of GSEditor generated by SuGaR, with our neural shader and mesh results. To qualitatively evaluate our meshes, we also used another mesh deformation-based editing method, TextDeformer~\cite{gao2023textdeformer}, by applying an adjusted prompt to our initial mesh. 
We also evaluated our method on the ShapeNet dataset~\cite{shapenet2015} to test its ability to transform everyday objects in Figure \ref{fig:shapenet_small}. We provide our Shapenet results and further qualitative comparison of models in DTU dataset in our supplementary materials.

\subsection{Implementation Details}
We conduct our experiments building on the NDS pipeline~\cite{nds} and use differentiable rendering pipeline by~\cite{nvdiffrast} on PyTorch~\cite{pytorch}. For initial mesh reconstruction and subsequent optimization, we follow the hyperparameters from the NDS and utilize their loss functions to optimize our editing process.

The diffusion model's effectiveness and the consistency of its updates depend on several hyperparameters. For the initial noise calculation we set our latent weight $\lambda=0.5$ and \([t_{min}, t_{max}]=[0,02, 0.98]\). Our denoising process is always done in 10 steps. The guidance scale for the text prompt is set at \(s_T = 7.5\), while the conditioning scales for normal maps and rendered images are \(s_I = 0.8\) and \(s_n = 0.2\), respectively. These parameters determine the influence of each component in the denoising process. We generate a image and update our dataset with the modified image every 10 iterations. Most of the results shown in the paper follow these initial parameters, changing these parameters can add variation to the result. We train our method for a maximum of 8k iterations which takes around 15 minutes on a single NVIDIA A40. Further training could result in additional shape alteration and the inclusion of extra features.

\begin{figure*}
   \centering
   \includegraphics[width=2.0\columnwidth,]{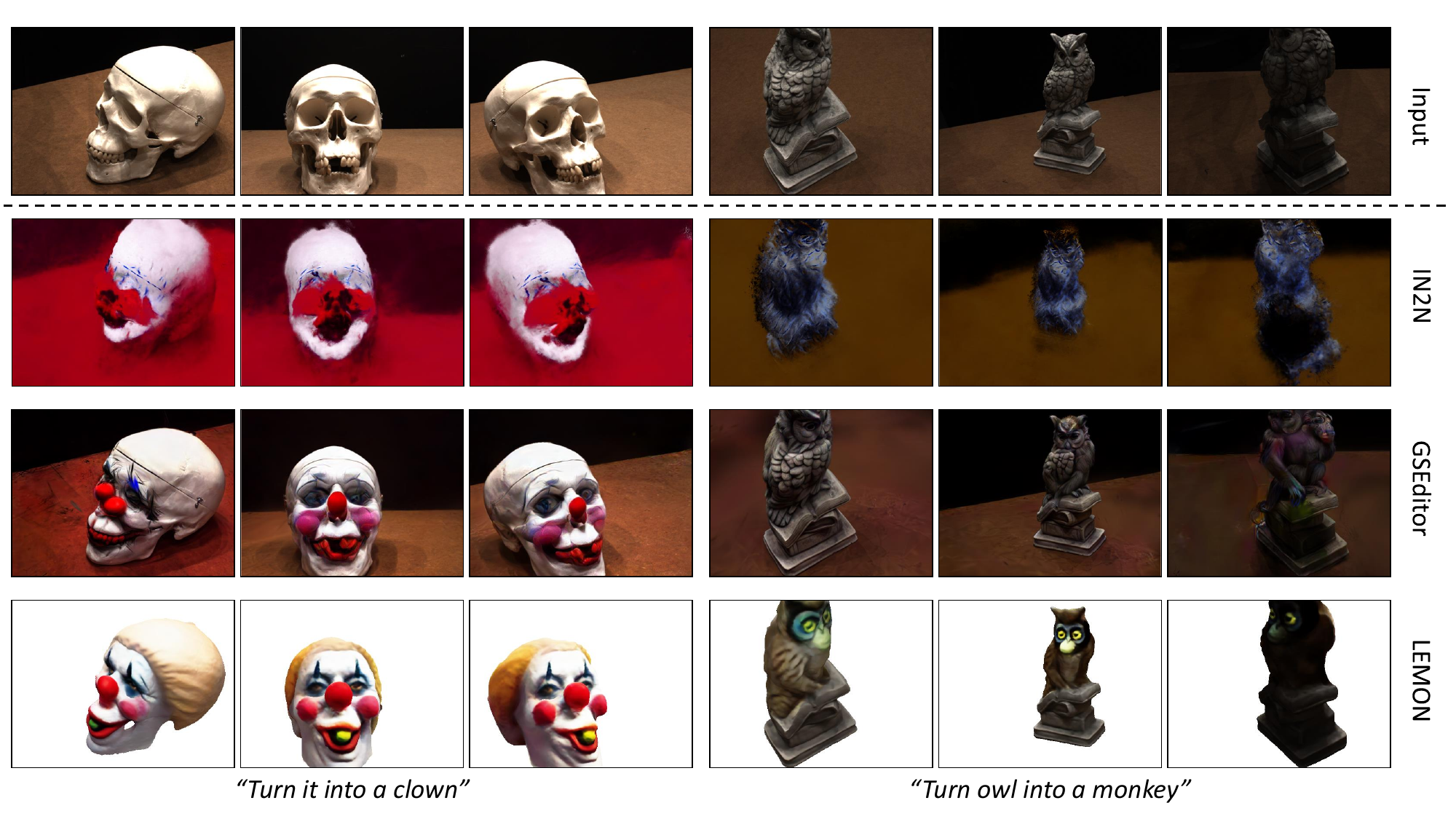} % adjust width as needed
 \caption{Qualitative consistency results.}
 \label{fig:qualitative_consistency} % optional label for cross-references
 \end{figure*}
\subsection{Qualitative Results}

%In Figure \ref{fig:shapenet_results}, we present our results on the ShapeNet dataset~\cite{shapenet2015}. Each row represents the initial ShapeNet object, and each column shows the text prompt given to our model, with phrases like "Turn it into {object}". The diagonals display the ground truth. The results demonstrate that our method can be used for object transformation. Minimal changes are observed in flat surfaces, attributed to their limited geometric characteristics. Notably, the transformation into a sofa yielded particularly favorable results, likely because sofas tend to have more distinct geometric characteristics compared to other objects.

In Figure \ref{fig:dtu_results}, we present our main comparison on the DTU dataset~\cite{DTU}. Each row represents a specific editing case with its corresponding text prompt listed below, while each column shows the output from a different method. In the case of mesh editing, our method retains the original mesh's geometric features while incorporating new refined details based on the text prompt. While TextDeformer achieves decent results it struggles to preserve the original structure, often leading to distortions like the horizontal compression of the skull shown in the first row. Even in seemingly successful cases like the second row, TextDeformer tends to overedit the mesh, whereas our method maintains the basic structure and adds specific details in line with the prompt. 

In the case of rendering, while Instruct-NeRF2NeRF achieves decent rendering results for simple edits like "Put in plate armor," it may fall short when dealing with editing a geometrically complex figure like the skull in the first row. Even in its most successful outcomes, Instruct-NeRF2NeRF seems to "color" the object rather than adding new geometrical features. GaussianEditor produces high-quality renderings of the edited object, however, it sometimes fails to generate the requested edits and shows less variation in the results. Unlike other two methods, our method maintains a concurrent relationship between mesh deformation and neural shader. This allows LEMON to adapt to significant structural changes while preserving the geometrical and shading characteristics of the object.As a result, our rendered images can achieve natural reflective effects, as seen in the first and second rows, demonstrating LEMON's effectiveness in both rendering and mesh editing.

We also compare our shader's consistency with other rendering methods in Figure \ref{fig:qualitative_consistency}. As seen in the figure, other methods tend to show more inconsistency when there is a drift in camera motion. Although our shader may not produce the highest quality renderings, it is more consistent than the other methods. We believe this is because the mesh and neural shader optimization processes are intertwined, providing geometric consistency to each other.
 % use [h] to keep the figure in place or [t], [b], [p] for top, bottom, or a separate page
 % use [h] to keep the figure in place or [t], [b], [p] for top, bottom, or a separate page
%\begin{figure*}
%   \centering
%   \includegraphics[width=0.85\textwidth,]{figures/ShapeNet-Figures-new.pdf} % adjust width as needed
 %\caption{ShapeNet editing results with shaded image and normal map of the edited mesh. We take multi-view images from a ShapeNet object and give the prompt "\textit{Turn it into }\{\textit{object}\}." Diagonals correspond to the image and normal of the ground truth object.}
 %\label{fig:shapenet_results} % optional label for cross-references
 %\end{figure*}

\begin{table}
\centering
\resizebox{\columnwidth}{!}{%
\begin{tabular}{|l|c|c|c|}
\hline
Method & Time $\downarrow$ & Memory $\downarrow$ & CLIP Similarity $\uparrow$ \\ 
       & (Mins.)          & (Peak GBs)          &                            \\ \hline\hline
Instruct-NeRF2NeRF~\cite{Haque_2023_ICCV} & $\sim$42 & $10.7$ & $0.1118$ \\
+Poisson Reconstruction~\cite{kazhdan2006poisson} & $\sim$ 3 & 6.0 &  \\ \hline
GaussianEditor~\cite{chen2024gaussianeditor} & $\sim$10 & 9.7 & $0.1262$ \\
+SuGaR~\cite{guedon2024sugar} & $\sim$45 & {6.5}  &  \\ \hline
LEMON(Ours) & $\sim$\textbf{15} & \textbf{6.7} & \textbf{0.2044} \\ \hline
\end{tabular}%
}
\caption{Quantitative results of our method.}
\label{tab:quantative}
\end{table}

\subsection{Quantitative Results}
Since editing is a subjective task we rely on our qualitative evaluation more. However, we also apply  CLIP Directional Similarity, introduced in StyleGAN-Nada~\cite{gal2021stylegannada} to measure the cosine similarity between the distance between pairs of images and the distance between pairs of captions accompanying the images. We evaluate on all views of the object dataset  We show our result in Table \ref{tab:quantative}, along with the time spent and the GPU memory consumption during the editing process.

In Table \ref{tab:quantative}, we see that our method outperforms Instruct-NeRF2NeRF and GaussianEditor in CLIP Directional Similarity. The higher CLIP Directional Similarity score indicates that our method more accurately follows the edit requests in the prompt. Since our pipeline simultaneously extracts the mesh and neural shader, we also consider the memory and time consumed in mesh extraction for the other methods. We outperform Instruct-NeRF2NeRF even without considering the mesh extraction time. Although GaussianEditor is faster, extracting meshes with SuGar takes considerably more time. Therefore, we believe our method strikes a good balance, achieving mesh and view synthesis together faster than all the other methods.

\begin{figure}
  \centering
  \includegraphics[width=\columnwidth,  ]{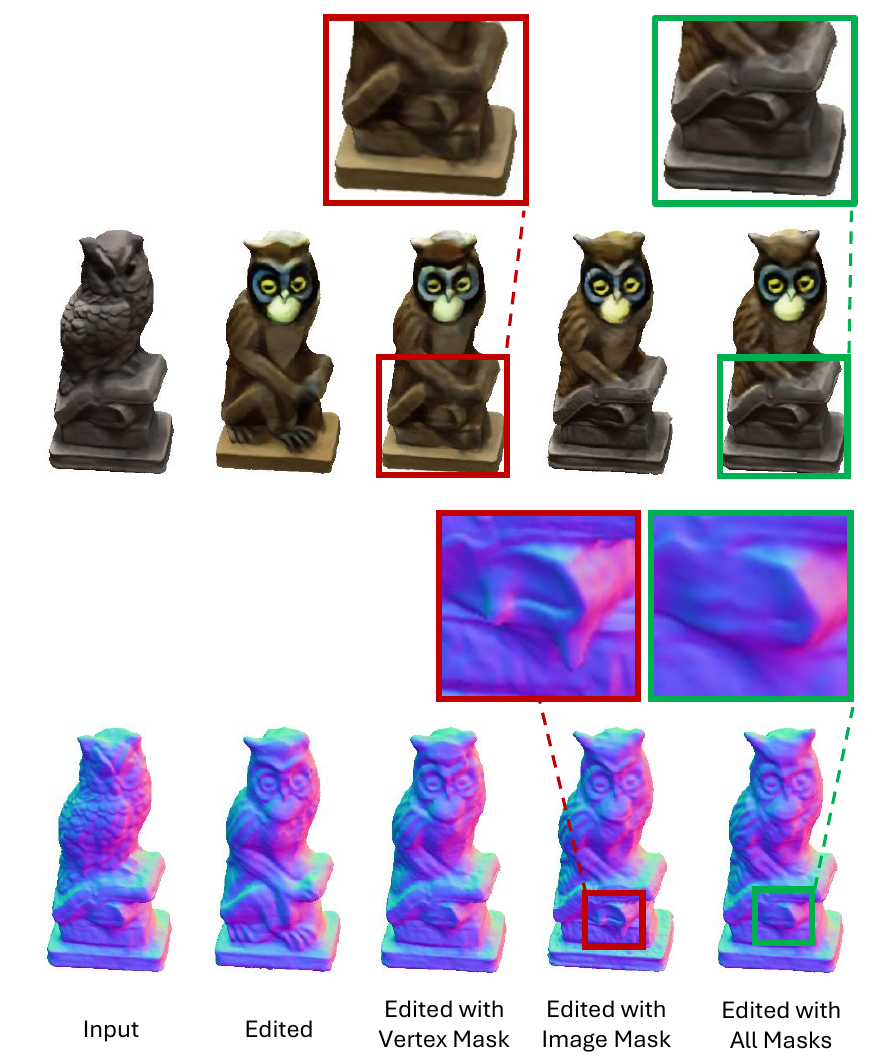} % adjust width as needed
\caption{Ablation studies of the masking.}
\label{fig:ablation_mask} % optional label for cross-references
\end{figure}

\subsection{Ablation Studies}

\begin{figure}
  \centering
  \includegraphics[width=\columnwidth, ]{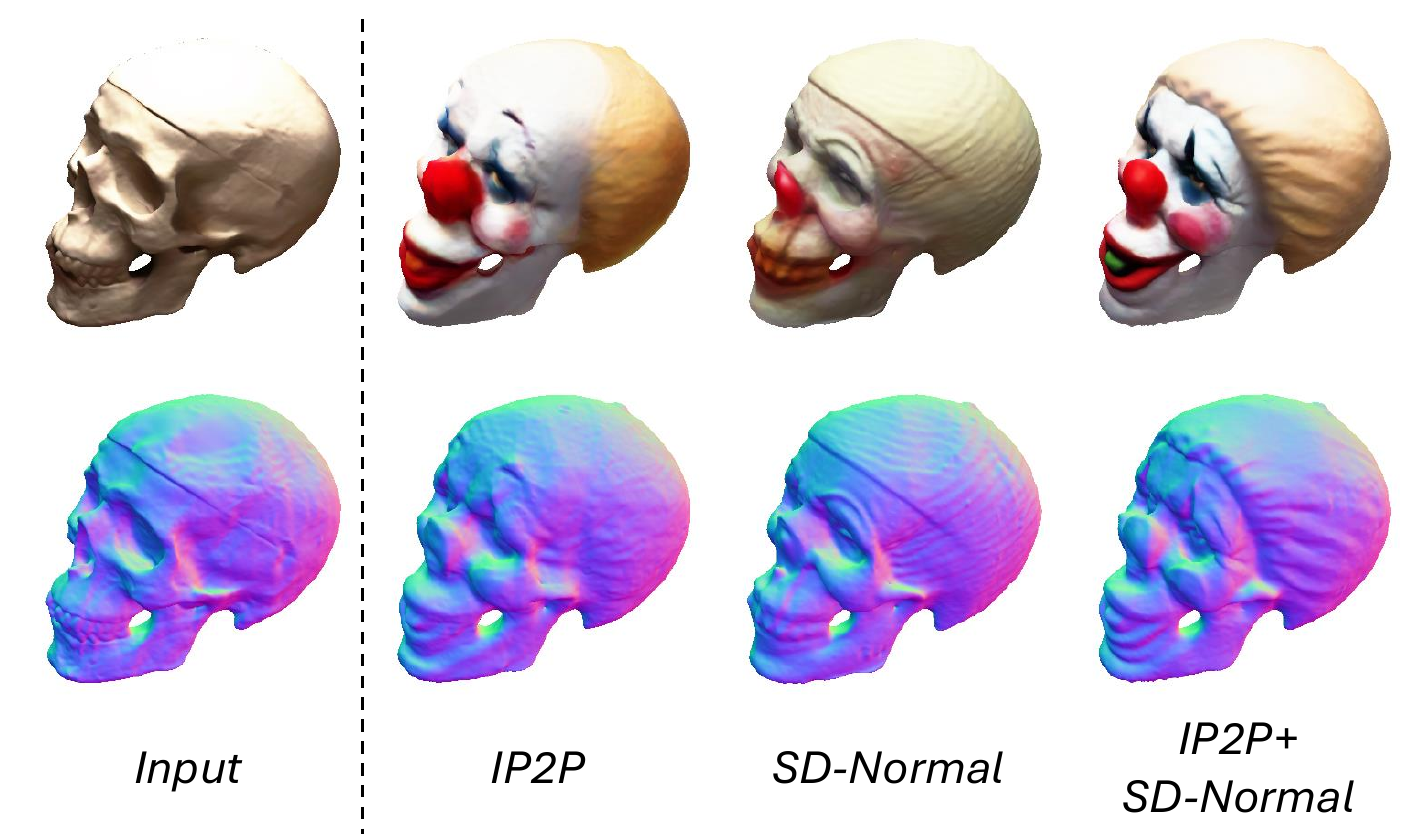} % adjust width as needed
\caption{Ablation studies of impact of text-to-image diffusion models on the mesh and neural shader.}
\label{fig:ablation_diffusion} % optional label for cross-references
\end{figure}

\textbf{\hspace{1em}Diffusion Model:} The diffusion model plays a key role in our editing process by generating our modifications. As shown in Figure \ref{fig:ablation_diffusion}, while Instruct-Pix2Pix introduces variation to the meshes, it also loses key details, such as the line on the skull. In contrast, ControlNet's normal model preserves the geometric details of the shape but provides less variety, making fewer changes. To achieve the best of both worlds, we connect the two models through a Multi-ControlNet pipeline.

\textbf{Vertice and Image Masking:} Vertex and image masking processes are the important contributions to our localized mesh editing. In Figure \ref{fig:ablation_mask}, we present a qualitative comparison of our masks on the owl example. Our vertex masking allows us to avoid unnecessary topological changes in the mesh, while image masking during the editing process avoids reduntant coloring preserving original shading of the input in the process.

\section{Conclusion and Future Work}
Even though our method is effective for editing meshes, it inherits many of the limitations associated with the methods that we used. Neural deferred shading relies on object masks to reconstruct the object, with appearance loss functioning only on the masked regions. This restricts the diffusion model, making it more challenging to add new objects to the mesh.  We believe this issue could be addressed by incorporating inpainting, similar to the approach used in GSEditor~\cite{chen2024gaussianeditor}.

\section{Acknowledgements}
The authors acknowledge the financial support  by the Bavarian Ministry of Economic Affairs, Regional Development and Energy as part of the project 6G Future Lab Bavaria.

%Moreover, updating the vertices of an initial mesh,  does not accommodate changes in topology. This limitation could lead to challenges when attempting to edit meshes that require topological modifications, such as adding or removing parts of the mesh. Future work could explore methods to integrate topology-aware mesh editing to overcome this limitation,

%In addition to mesh editing, our neural shader also demonstrates satisfactory performance. However, as in NeRFs, it can also produce view-dependent artifacts. To improve this, we aim to increase the consistency of our neural shader, which we believe will lead to better mesh editing results. Currently, our method employs pre-trained ControlNet variants, which can limit flexibility due to their pre-set conditions. For future improvements, we plan to train a ControlNet model conditioned on mesh features, and if feasible, train it concurrently with mesh deformation. This approach could lead to more consistent edits across the triangulated mesh and improve the neural shader's performance.

%\section{Conclusion}

\bibliographystyle{ieeenat_fullname}
\bibliography{main}

\clearpage
\setcounter{page}{1}
\maketitlesupplementary

\section{Appendix}
\label{sec:appendix}
\subsection{Additional Qualitative Results}
\begin{figure*}
  \centering
  \includegraphics[width=2\columnwidth, ]{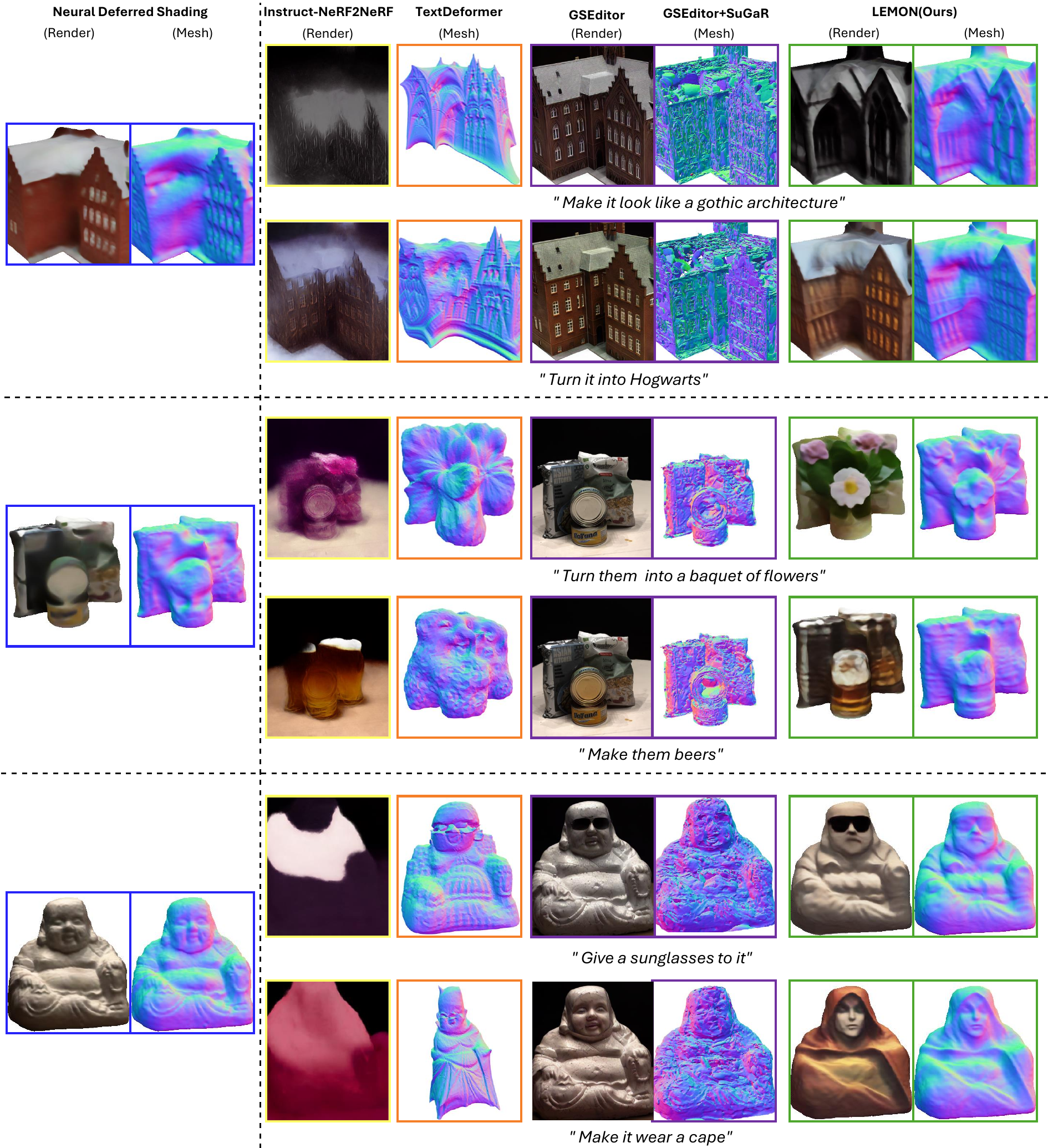} % adjust width as needed
\caption{More editing results on the DTU dataset~\cite{DTU}.}
\label{fig:supp1}
\end{figure*}

\begin{figure*}
  \centering
  \includegraphics[width=2\columnwidth, ]{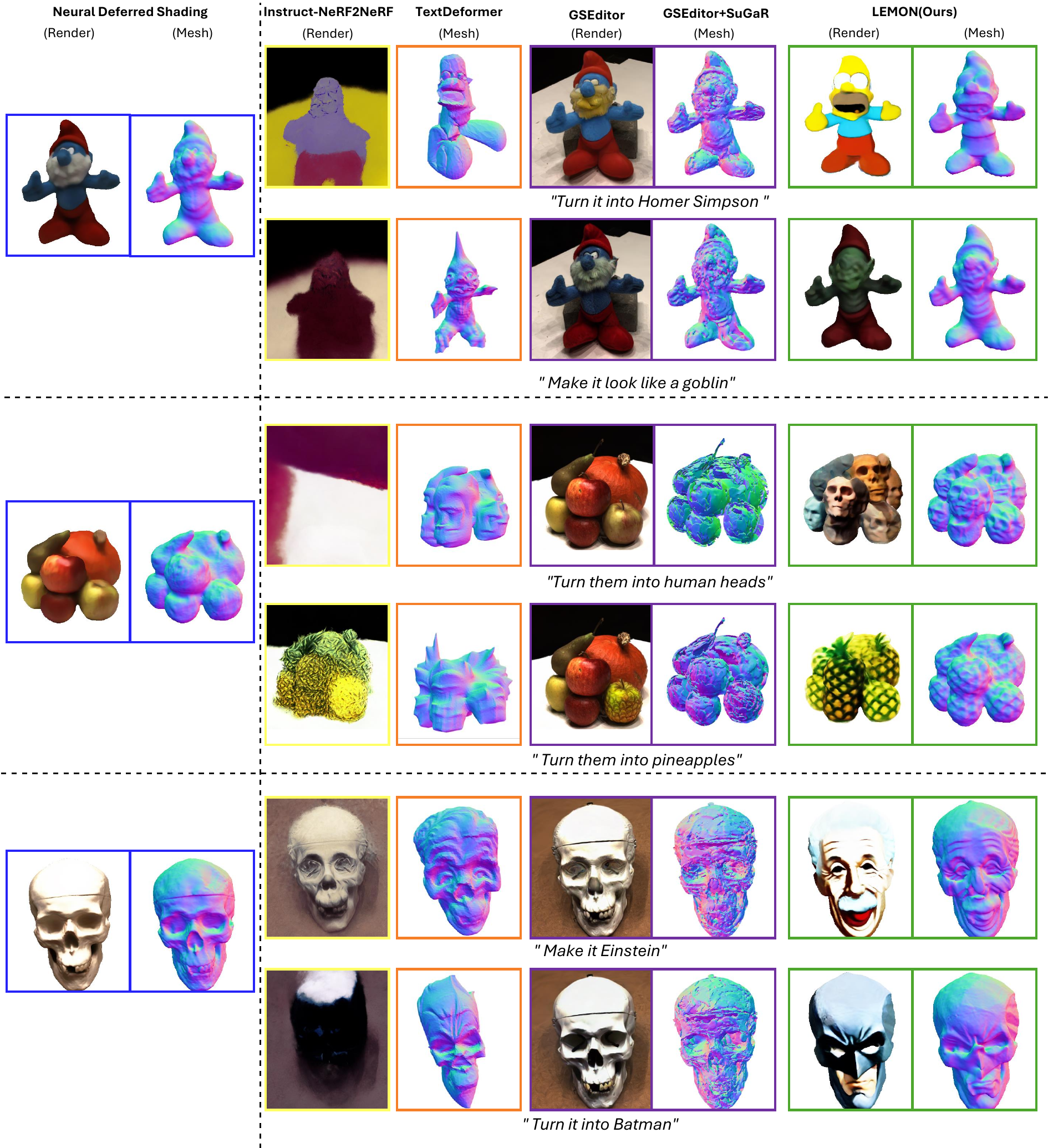} % adjust width as needed
\caption{More editing results on the DTU dataset~\cite{DTU}.}
\label{fig:supp2}
\end{figure*}

\begin{figure*}
  \centering
  \includegraphics[width=2\columnwidth, ]{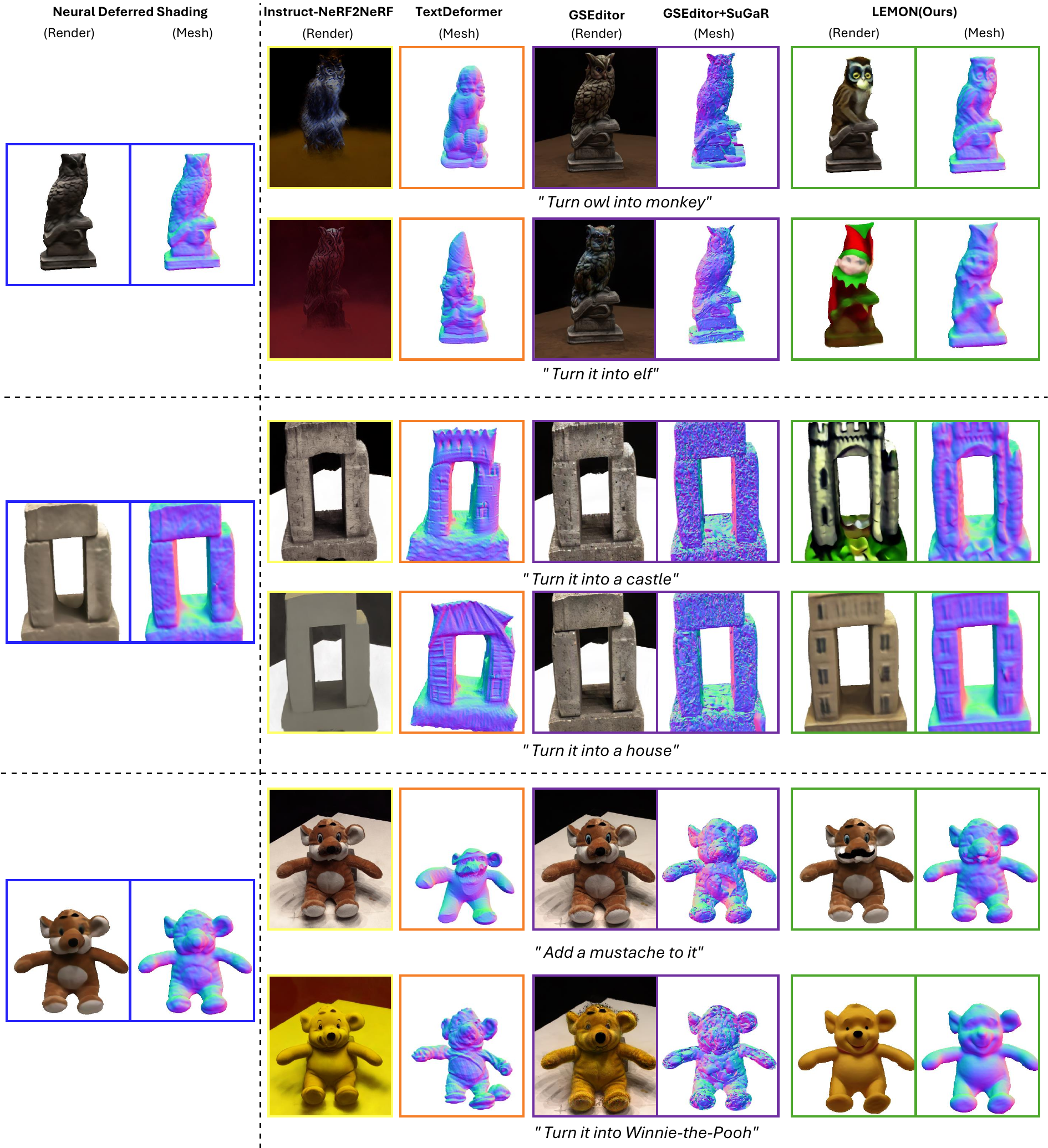} % adjust width as needed
\caption{More editing results on the DTU dataset~\cite{DTU}.}
\label{fig:supp3}
\end{figure*}

We present additional qualitative results in this section, extending those presented in Figure \ref{fig:dtu_results} of the main paper. Figures \ref{fig:supp1},\ref{fig:supp2},\ref{fig:supp3} shows further experiments, with each column corresponding to the methods used and each row representing the object. Text instructions that provided in experiments are below each row. It's important to note that results may vary depending on the hyperparameters of these models.

\begin{figure*}
  \centering
  \includegraphics[width=2\columnwidth, ]{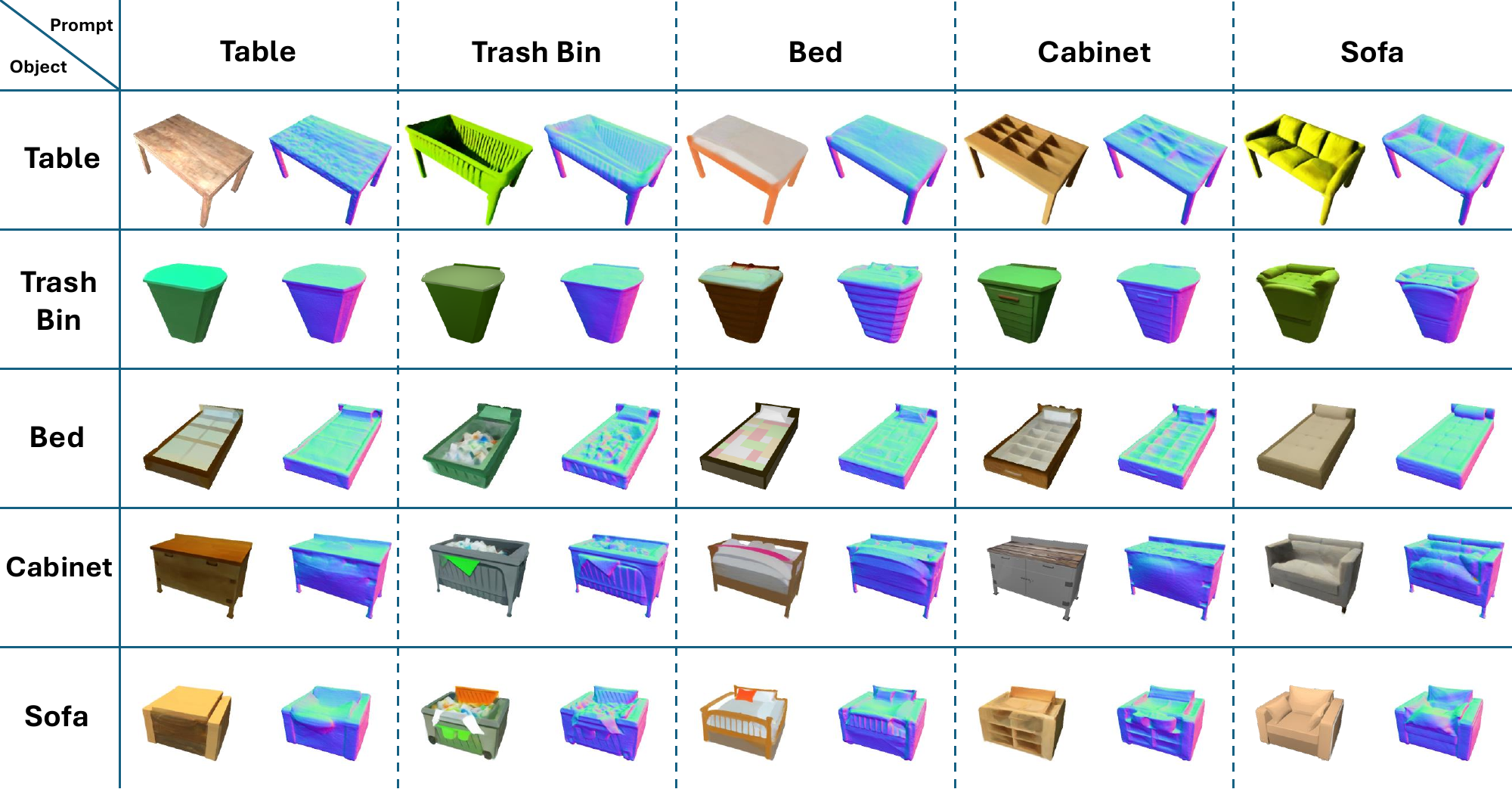} % adjust width as needed
\caption{Extended ShapeNet editing results with shaded image and normal map of the edited mesh. We take multi-view images from a ShapeNet object and give the prompt "\textit{Turn it into }\{\textit{object}\}." Diagonals correspond to the image and normal of the ground truth object.}
\label{fig:supp_shape}
\end{figure*}

In Figure \ref{fig:supp_shape}, we present our extended results on the ShapeNet dataset~\cite{shapenet2015}. Each row represents the initial ShapeNet object, and each column shows the text prompt given to our model, with phrases like "Turn it into {object}". The diagonals display the ground truth. The results demonstrate that our method can be used for object transformation. Minimal changes are observed in flat surfaces, attributed to their limited geometric characteristics. Notably, the transformation into a sofa yielded particularly favorable results, likely because sofas tend to have more distinct geometric characteristics compared to other objects.

\subsection{Code and Video}
We attach our code to our supplementary material. The results presented in our paper can be replicated by following the instructions provided in the code folder. We also provide timelapse videos of the training process and a video demonstrating the consistency of the neural shader.

% use [h] to keep the figure in place or [t], [b], [p] for top, bottom, or a separate page

 % use [h] to keep the figure in place or [t], [b], [p] for top, bottom, or a separate page

 % use [h] to keep the figure in place or [t], [b], [p] for top, bottom, or a separate page

 % use [h] to keep the figure in place or [t], [b], [p] for top, bottom, or a separate page

\end{document}